\documentclass[10pt,twocolumn,letterpaper]{article}

\usepackage{cvpr}
\usepackage{times}
\usepackage{epsfig}
\usepackage{graphicx}
\usepackage{amsmath}
\usepackage{amssymb}
\usepackage{booktabs}
\usepackage{float}
\usepackage{tabularx}
\usepackage{verbatim}
\usepackage{color}
\usepackage{url}

\usepackage[breaklinks=true,bookmarks=false]{hyperref}

\cvprfinalcopy 

\def\cvprPaperID{2531} 

\begin{document}

\title{CentripetalNet: Pursuing High-quality Keypoint Pairs for Object Detection}

 \author{Zhiwei Dong\textsuperscript{\rm 1,2}\quad Guoxuan Li\textsuperscript{\rm 3} \quad Yue Liao\textsuperscript{\rm 4}\quad Fei Wang\textsuperscript{\rm 2}\thanks{Corresponding author}\quad Pengju Ren\textsuperscript{\rm 1}\quad Chen Qian\textsuperscript{\rm 2}\\
\large\textsuperscript{\rm 1} Institute of Artificial Intelligence and Robotics, Xi’an Jiaotong University \\\quad\textsuperscript{\rm 2}SenseTime Research\quad\textsuperscript{\rm 3}University of Chinese Academy of Sciences \quad\textsuperscript{\rm 4}Beihang University\\
\tt\small kivee@foxmail.com; liguoxuan18@mails.ucas.ac.cn; liaoyue.ai@gmail.com; \\ \tt\small\{wangfei,\ qianchen\}@sensetime.com; \tt\small pengjuren@xjtu.edu.cn}

\maketitle
\thispagestyle{empty}
\pagestyle{empty}

\begin{abstract}
Keypoint-based detectors have achieved pretty-well performance. However, incorrect keypoint matching is still widespread and greatly affects the performance of the detector. In this paper, we propose CentripetalNet which uses centripetal shift to pair corner keypoints from the same instance. CentripetalNet predicts the position and the centripetal shift of the corner points and matches corners whose shifted results are aligned. Combining position information, our approach matches corner points more accurately than the conventional embedding approaches do. Corner pooling extracts information inside the bounding boxes onto the border. To make this information more aware at the corners, we design a cross-star deformable convolution network to conduct feature adaption. Furthermore, we explore instance segmentation on anchor-free detectors by equipping our CentripetalNet with a mask prediction module. On MS-COCO test-dev, our CentripetalNet not only outperforms all existing anchor-free detectors with an $AP$ of $48.0\%$ but also achieves comparable performance to the state-of-the-art instance segmentation approaches with a $40.2\%$ $Mask AP$. Code will be available at \url{https://github.com/KiveeDong/CentripetalNet}.
\end{abstract}

\section{Introduction}

Object detection is a fundamental topic in various applications of computer vision, such as automatic driving, mobile entertainment, and video surveillance. 
It is challenging in large appearance variance caused by scale, deformation, and occlusion. 
With the development of deep learning, object detection has achieved great progress~\cite{girshick2014rich, girshick2015fast, ren2015faster, redmon2016you, Liu2016SSD, Lin2017Feature, He2017Mask, Lin2017Focal, Cai2018Cascade, law2018cornernet}. 
The anchor-based methods~\cite{girshick2015fast,ren2015faster,Liu2016SSD} have led the fashion in the past few years, but it is difficult to manually design a set of suitable anchors. Additionally, the anchor-based methods suffer from the significant imbalance between negative and positive anchor boxes. 
To improve it, the CornerNet~\cite{law2018cornernet} proposes a novel method to represent a bounding box as a pair of corners, i.e, top-left corner and bottom-right corner. 
Based on this idea, lots of corner-based methods~\cite{law2018cornernet,duan2019centernet} have emerged. 
The corner-based detection framework has been leading the new trends in the object detection area gradually. 
The corner-based detection framework can be divided into two steps including corner points prediction and corner matching. In this paper, we concentrate on the second step.
\begin{figure}[t]
  \centering
  \includegraphics[width=1\linewidth]{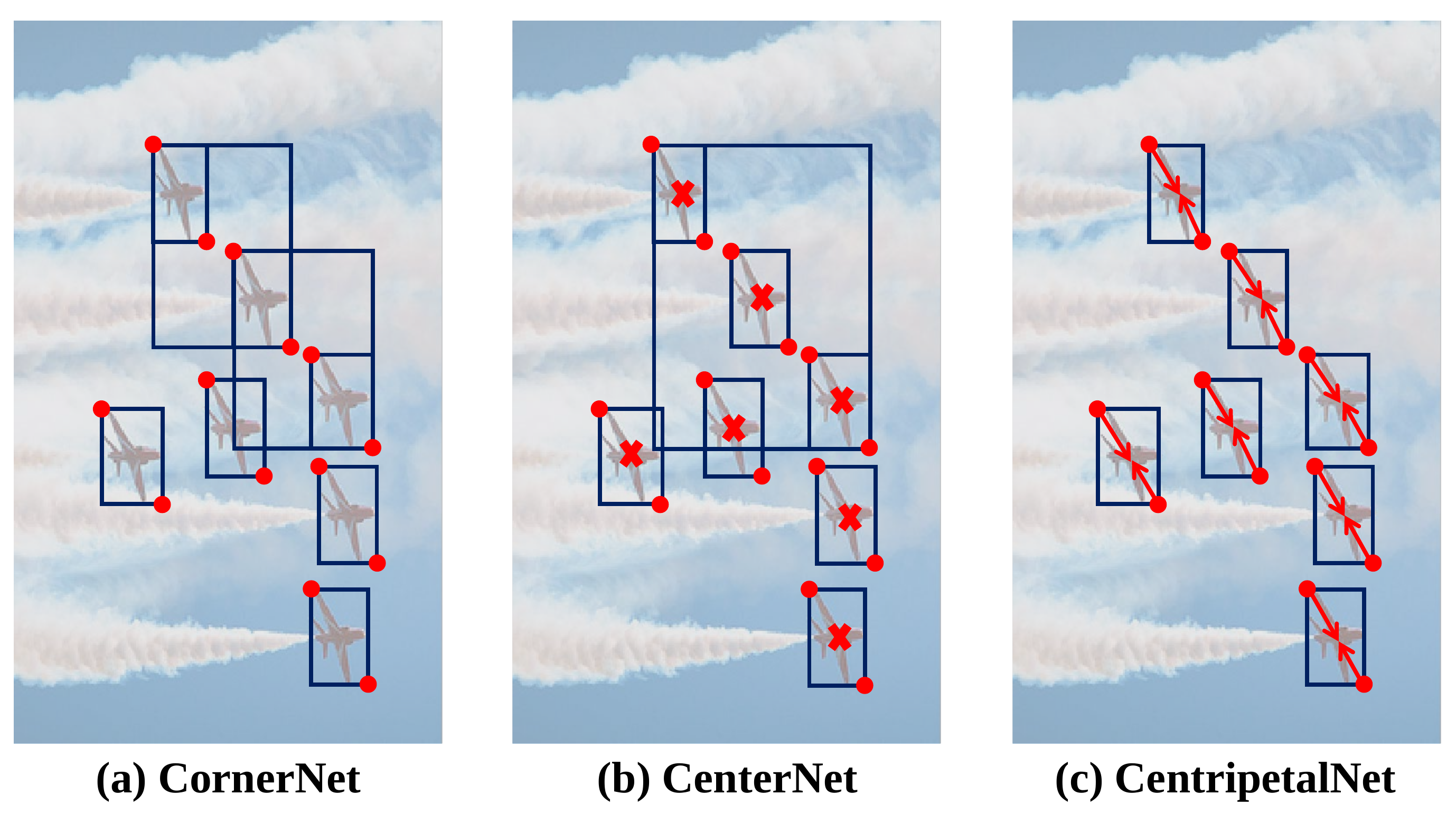}
  \caption{(a) CornerNet generates some false corner pairs because of similar embeddings caused by similar appearance. (b) CenterNet removes some false corner pairs through center prediction, but it naturally can not handle some dense situation. (c) CentripetalNet avoids the drawbacks of CornerNet and CenterNet.}
  \label{fig:motiv}
  \vspace{-2mm}
\end{figure}

The conventional methods~\cite{law2018cornernet,duan2019centernet} mainly use an associative embedding method to pair corners, where the network is required to learn an additional embedding for each corner to identify whether two corners belong to the same bounding-box. 
In this manner, if two corners are from the same box, they will have a similar embedding, otherwise, their embeddings will be quite different. Associative embedding-based detectors have achieved pretty-well performance in object detection, but they also have some limitations. 
Firstly, the training process employs push and pull loss to learn the embedding of each point. Push loss will be calculated between points that do not belong to the same object to push them away from each other. While the pull loss is only considered between points from the same object. 
Thus, during training, the network is actually trained to find the unique matching point within all potential points of the diagonal. 
It is highly sensitive to outliers and the training difficulty will increase dramatically when there are multiple similar objects in one training sample. Secondly, the embedding prediction is based on the appearance feature without using position information, thus as shown in Figure~\ref{fig:motiv}, if two objects have a similar appearance, the network tends to predict the similar embeddings for them even if they are far apart. 

Based on the above considerations, we propose a novel CentripetalNet using a corner matching method based on centripetal shift, along with a cross-star deformable convolution module for better prediction of centripetal shift. 
Given a pair of corners, we define a 2-D vector, i.e., centripetal shift, for each corner, where the centripetal shift encodes the spatial offset from the corner to the center point of the box. 
In this way, each corner can generate a center point based on the centripetal shift, thus if two corners belong to the same bounding-box, the center points generated by them should be close. 
The quality of the match may be represented by the distance between two centers and the geometric center of this match. 
%
%
Combined with position information of each corner point, the method is robust to outliers compared to the associative embedding approach. 
%
Furthermore, we propose a novel component, namely cross-star deformable convolution, to learn not only a large receptive field but also the geometric structure of `cross star'. We observe that there are some `cross stars' in the feature map of the corner pooling output.

The border of the `cross star' contains context information of the object because corner pooling uses $max$ and $sum$ operations to extend the location information of the object to the corner along the `cross star' border. Thus, we embed the object geometric and location information into the offset field of the deformable convolution explicitly.
%
Equipped with the centripetal shift and cross-star deformable convolution, our model has achieved a significant performance gain compared to CornerNet, from $42.1\%$ AP to $47.8\%$ AP on MS-COCO test-dev2017. 
Moreover, motivated by the benefits of multi-task learning in object detection, we first add instance mask branch to further improve the accuracy. 
We apply the RoIAlign to pool features from a group of predicted regions of interests(RoIs) and feed the pooled features into a mask head to generate the final segmentation prediction. 
%
%
To demonstrate the effectiveness of the proposed CentripetalNet, we evaluate the method on the challenging MS-COCO benchmark~\cite{lin2014microsoft}. 
CentripetalNet not only outperforms all existing anchor-free detectors with an AP of 48.0\% but also achieves comparable performance with the state-of-the-art instance segmentation methods on MS-COCO test-dev.
%
\section{Related Work}

\noindent\textbf{Anchor-based Approach:} 
Anchor-based detectors set anchor boxes in each position of the feature map. The network predicts the probability of having objects in each anchor box and adjusts the size of the anchor boxes to match the object. Generally, anchor-based methods can be divided into two types, namely two-stage methods and single-stage methods. 

Two-stage methods are derived from R-CNN series of methods~\cite{girshick2014rich, he2015spatial, girshick2015fast} which first extract RoIs using a selective search method~\cite{uijlings2013selective} then classify and regress them. Faster R-CNN~\cite{ren2015faster} employs a region proposal network(RPN) to generate RoIs by modifying preset anchor boxes. Mask R-CNN~\cite{He2017Mask} replaces the RoIPool layer with the RoIAlign layer using bilinear interpolation. Its mask head uses a top-down method to obtain instance segmentations. 

Without extracting RoIs, one-stage methods directly classify and regress the preset anchor boxes. SSD~\cite{Liu2016SSD} utilizes features maps from multiple different convolution layers to classify and regress anchor boxes with different strides. Compared with YOLO~\cite{redmon2016you}, YOLOv2~\cite{redmon2017yolo9000} uses preset anchors. However, the above methods are bothered by the imbalance between negative and positive samples. RetinaNet~\cite{Lin2017Focal} uses focal loss to mitigate classification imbalance problem. RefineDet~\cite{Zhang2018Single} refines the FPN structure by introducing the anchor refinement module to filter and eliminate negative samples. 

Other works cooperating with anchor-based detectors are proposed to deal with different issues, such as improving anchor selection procedure~\cite{Wang2019Region}, refining feature learning process~\cite{Zhu2019Feature, li2019scale}, optimizing location prediction method~\cite{lu2019grid}, and improving the loss function~\cite{rezatofighi2019generalized, jiang2018acquisition}.

\begin{figure*}[htb]
  \centering
  \includegraphics[width=1\linewidth]{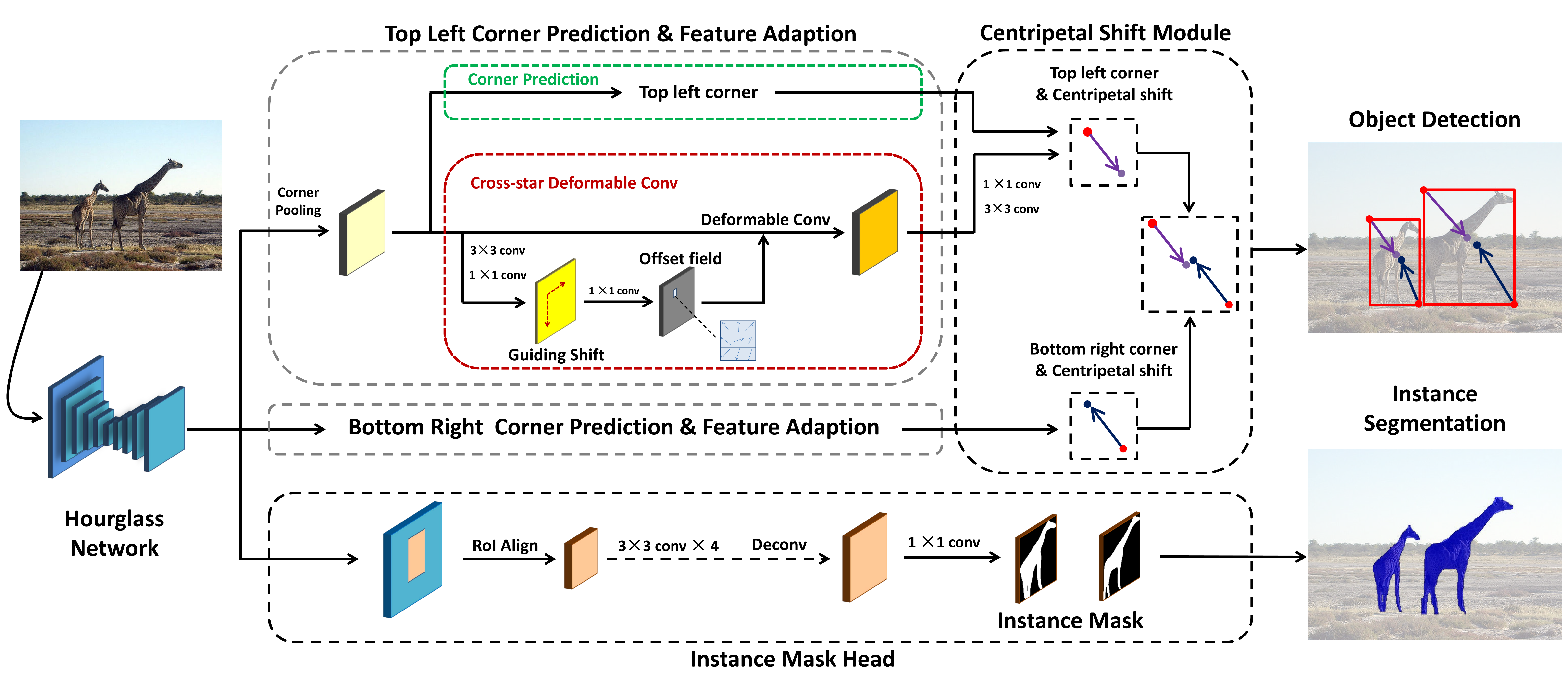}
 \caption{An overview of CentripetalNet. As the corner prediction and feature adaption of top-left corner and bottom-right corner are similar, we only draw top-left corner module for simplicity. Centripetal shift module gets predicted corners and adapted features, then it predicts the centripetal shift of each corner and performs corner matching based on the predicted corners and centripetal shifts. During matching, if the positions of the shifted corners are close enough, they form a bounding box with a high score.}
  \label{fig:f_work}
\end{figure*}

\vspace{1mm}
\noindent\textbf{Anchor-free Approach:}
For anchor-based methods, the shape of anchor boxes should be carefully designed to fit the target object. Compared to the anchor-based approach, anchor-free detectors no longer need to preset anchor boxes. Mainly two types of anchor-free detectors are proposed. 

The first type of detectors directly predict the center of an object. Yolov1~\cite{redmon2016you} predicts the size and shape of the object at the points near the center of the object. DenseBox~\cite{huang2015densebox} introduces a fully convolutional neural network framework to gain high efficiency. UnitBox~\cite{yu2016unitbox} uses IoU loss to regress the four bounds as a whole unit. Since the number of positive samples is relatively small, these detectors suffer from a quite low recall. To cope with this problem, FCOS~\cite{tian2019fcos} treats all the points inside the bounding box of the object as positive samples. It detects all the positive points and the distance from the point to the border of the bounding box. 

For the second type, detectors predict keypoints and group them to get bounding boxes. CornerNet~\cite{law2018cornernet} detects top-left and bottom-right corners of the object and embeds them into an abstract feature space. It matches corners of the same object by computing distance between embeddings of each pair of points. ExtremeNet~\cite{zhou2019bottom} detects the top-, left-, bottom-, rightmost, and center points of the object. Combined with Deep Extreme Cut~\cite{maninis2018deep}, the extreme points can be used for instance segmentation. These detectors need some specific grouping methods to obtain bounding boxes.  RepPoints~\cite{yang2019reppoints} uses deformable convolutional networks(DCN)~\cite{dai2017deformable} to get sets of points used to represent objects. The converting functions are carefully designed to convert point sets to bounding boxes. CenterNet~\cite{duan2019centernet} adds a center detection branch into CornerNet and largely improves the performance by center point validation. 

These methods usually achieve high recall with quite many false detections. The main challenge resides in the approach to match keypoints of the same object. In this work, we propose a centripetal shift which encodes the relationship between corners and gets their corresponding centers by predicted spatial information, thus we can build the connection between the top-left and bottom-right corners through their sharing center.

\section{CentripetalNet}

We first provide an overview of the approach. 
As shown in Figure~\ref{fig:f_work}, CentripetalNet consists of four modules, namely corner prediction module, centripetal shift module, cross-star deformable convolution, and instance mask head. 
We first generate corner candidates based on the CornerNet pipeline. 
With all the corner candidates, we then introduce a centripetal shift algorithm to pursue high-quality corner pairs and generate final predicted bounding boxes. 
%
Specifically, the centripetal shift module predicts the centripetal shifts of the corner points and matches corner pairs whose shifted results decoded from their locations and centripetal shifts are aligned. 
Then, we propose a novel cross-star deformable convolution, whose offset field is learned from the shifts from corners to their corresponding centers, to conduct feature adaption for enriching the visual features of the corner locations, which is important to improve the accuracy of the centripetal shift module. 
%
%
Finally, we add an instance mask module to further improve the detection performance and extend our method to the instance segmentation area. Our method takes the predicted bounding boxes of centripetal shift module as region proposals, uses RoIAlign to extract the region features and applies a small convolution network to predict the segmentation masks. Overall, our CentripetalNet is trained end-to-end and can inference with or without the instance segmentation module.

\begin{figure}[htb]
  \centering
  \includegraphics[width=1.0\linewidth]{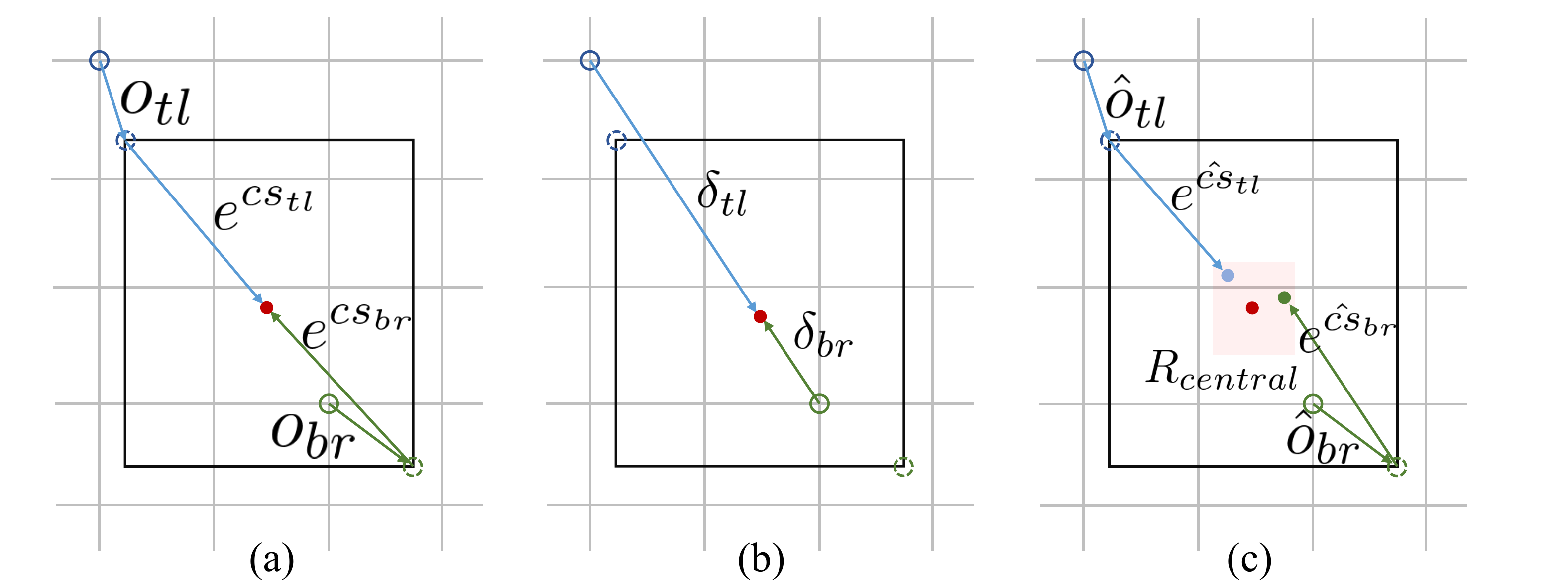}
  \caption{(a) When mapping the ground truth corner to the heatmap, local offset $O_{tl}$(or $O_{br}$) is used to compensate the precision loss as in~\cite{law2018cornernet}. (b) The guiding shift $\delta$ is the shift from ground truth corner on the heatmap to center of bounding box. (c) $R_{central}$ is the central region we use to match the corners.}
\label{fig:offset}
\end{figure}

\subsection{Centripetal Shift Module}
\noindent\textbf{Centripetal Shift.}
For $bbox^i = (tlx^i, tly^i, brx^i, bry^i)$, its geometric center is $(ctx^i, cty^i) = (\frac{tlx^i + brx^i}{2}, \frac{tly^i + bry^i}{2})$. We define the centripetal shifts for its top-left corner and bottom-right corner separately as
\begin{equation}
\begin{array}{l}
    cs_{tl}^i = (log(\frac{ctx^i - tlx^i}{s}), log(\frac{cty^i - tly^i}{s})) \vspace{2mm }\\
    
    cs_{br}^i = (log(\frac{brx^i - ctx^i}{s}), log(\frac{bry^i - cty^i}{s}))
\end{array}
\end{equation}
Here we use $log$ function to reduce the numerical range of centripetal shift and make the learning process easier.

During training, we apply smooth L1 loss at the locations of ground truth corners
\begin{equation}
    L_{cs} = \frac{1}{N}\sum^N_{k=1}[\mathcal{L}_1(cs^k_{tl}, \hat{cs}^k_{tl}) + \mathcal{L}_1(cs^k_{br}, \hat{cs}^k_{br})]
\end{equation}
where $\mathcal{L}_1$ is SmoothL1 loss and $N$ is the number of ground truths in a training sample.

\vspace{1mm}
\noindent\textbf{Corner Matching.}
To match the corners, we design a matching method using their centripetal shifts and their locations. It is intuitive and reasonable that a pair of corners belonging to the same bounding box should share the center of that box. As we can decode the corresponding center of a predicted corner from its location and centripetal shift, it is easy to compare whether the centers of a pair of corners are close enough and close to the center of the bounding box composed of the corner pair, as shown in Figure~\ref{fig:offset}(c). Motivated by the above observations, our method goes as follows. Once the corners are obtained from corner heatmaps and local offset feature maps, we group the corners that are of the same category and satisfy $tlx<brx\land tly<bry$ to construct predicted bounding boxes. For each bounding box $bbox^j$, we set its score as the geometric mean of its corners' scores, which are obtained by applying $softmax$ on predicted corner heatmaps.

Then, as shown in Figure~\ref{fig:offset} we define a central region for each bounding box as Equation~\ref{eql:r_ctr} to compare the proximity of decoded centers and the bounding box center.
\begin{equation}
\vspace{-1mm}
R_{central} = \{(x,y)|x\in[ctlx,cbrx], y\in[ctly,cbry]\}
\label{eql:r_ctr}
\end{equation}
and the corners of $R_{central}$ are computed as
\begin{equation}
\left\{
\begin{array}{l}
    ctlx = \frac{tlx + brx}{2} - \frac{brx - tlx}{2}\mu \vspace{1.5mm}\\
    ctly = \frac{tly + bry}{2} - \frac{bry - tly}{2}\mu \vspace{1.5mm}\\
    cbrx = \frac{tlx + brx}{2} + \frac{brx - tlx}{2}\mu \vspace{1.5mm}\\
    cbry = \frac{tly + bry}{2} + \frac{bry - tly}{2}\mu 
\end{array}
\right.
\end{equation}
where $0<\mu\le1$ indicates that  width and height of central region are $\mu$ times of the bounding box's width and height. With the centripetal shift, we can decode the center $(tl_{ctx}, tl_{cty})$ and $(br_{ctx}, br_{cty})$ for top-left corner and bottom-right corner separately.

Then we calculate the score weight $w^j$ for each predicted bounding box that satisfies $(tl^j_{ctx}, tl^j_{cty}){\in}R^j_{central}\land(br^j_{ctx}, br^j_{cty}){\in}R^j_{central} $ as follows
\begin{equation}
    w^j = e^{-\frac{|br^j_{ctx} - tl^j_{ctx}||br^j_{cty} - tl^j_{cty}|}{(cbrx^j - ctlx^j)(cbry^j - ctly^j)}}
\end{equation}
which means that the regressed centers are closer, the predicted box has a higher scoring weight. For other bounding boxes, we set $w^j = 0$. Finally we can re-score the predicted bounding boxes by multiplying the score weights.

\subsection{Cross-star Deformable Convolution}
\begin{figure}[htb]
  \centering
  \includegraphics[width=1\linewidth]{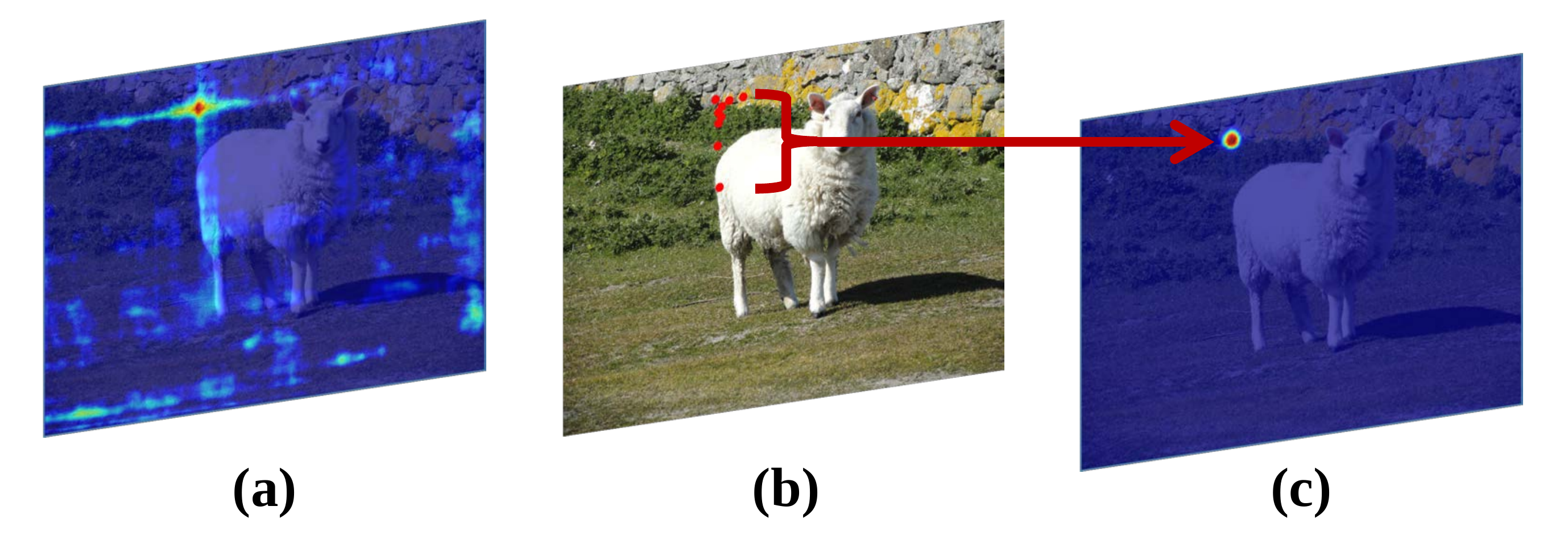}
  \caption{(a) `Cross star' caused by corner pooling. (b) The sampling points of the cross-star deformable convolution at the corner. (c) Top-left corner heatmap from corner prediciton module.}
  \label{fig:cross_star}
\end{figure}

Due to corner pooling, there are some `cross stars' in the feature map as shown in Figure~\ref{fig:cross_star}(a). The border of the `cross star' maintains abundant context information of the object because corner pooling uses $max$ and $sum$ operations to extend the location information of the object to the corner along the `cross star' border. To capture the context information on `cross star', not only a large receptive field is required, but also the geometric structure of `cross star' should be learned. Following the above intuition, we proposed the cross-star deformable convolution, a novel convolution operation to enhance the visual features at corners.

Our proposed cross-star deformable convolution is depicted in Figure~\ref{fig:f_work}. Firstly, we feed the feature map of the corner pooling into the cross-star deformable convolution module. 
%
To learn the geometric structure of `cross star' for deformable convolution, we can use the size of the corresponding object to guide the offset field branch explicitly, as we find that the shape of the `cross star' relates to the shape of the bounding box. However, take the top-left corner as an example, it is natural that they should pay less attention to the top-left part of the `cross star', as there is more useless information outside the object. So we embed a guiding shift, the shift from corner to center as shown in Figure~\ref{fig:offset}(b), which contains both shape and direction information, to the offset field branch. Specifically, the offset field is carried out on three convolution layers. The first two convolution layers embed the corner pooling output into the feature map, which is supervised by the following loss: 
\begin{equation}
  L_{\delta} = \frac{1}{N}\sum^N_{k=1}[L_1(\delta_{tl}, \hat{\delta}_{tl}) + L_1(\delta_{br}, \hat{\delta}_{br})]
\end{equation}
where $\delta$ means the guiding shift and is defined as
\begin{equation}
\label{eq_offsetfield_loss}
    {\delta}^i_{tl} = (\frac{ctx^i}{s} - {\lfloor}\frac{tlx^i}{s}{\rfloor}, \frac{cty^i}{s} - {\lfloor}\frac{tly^i}{s}{\rfloor})
\end{equation}
The second convolution layer maps the above feature into the offset field, which contains the context and geometric information explicitly. 
By visualizing the learned offset field as shown in Figure~\ref{fig:dcn_visual}c, our cross-star deformable convolution can efficiently learn the geometric information of `cross star' and extract information of `cross star' border.

\subsection{Instance Mask Head}
To get the instance segmentation mask, we treat the detection results before soft-NMS as region proposals and use a fully convolutional neural network to predict the mask on top of them. To make sure that the detection module could produce proposals, we first pretrain CentripetalNet for a few epochs.
We select top $k$ scored proposals and perform RoIAlign on top of the feature map from the backbone network to get their features. We set the size of RoIAlign to $14\times14$ and predict a mask of $28\times28$. 

After getting the features from RoIs, we apply four consecutive $3\times3$ convolution layers, then use a transposed convolution layer to upsample the feature map to a $28\times28$ mask map $\hat{m}$. During training, we apply cross entropy loss for each region proposal
\begin{equation}
    L_{mask} = \frac{1}{N}\sum^N_{k=1}CE(m_i, \hat{m_i})
    \vspace{-2mm}
\end{equation}

\begin{table*}[ht]
\begin{center}
\setlength{\tabcolsep}{3.8mm}{

\small
\begin{tabular}{l|l|ccc|ccc}
\hline
Method & Backbone & $AP$ & $AP_{50}$ & $AP_{75}$ & $AP_{S}$ & $AP_{M}$ & $AP_{L}$ \\
\hline\hline
\textbf{Two-stage:} & & & & & & & \\
Faster R-CNN w/FPN~\cite{Lin2017Feature} & ResNet-101~\cite{he2016deep} & 36.2 & 59.1 & 39.0 & 18.2 & 39.0 & 48.2  \\
Mask R-CNN~\cite{He2017Mask} & ResNeXt-101 & 39.8 & 62.3 & 43.4 & 22.1 & 43.2 & 51.2 \\
HTC~\cite{chen2019hybrid} & ResNeXt-101 & 47.1 & 63.9 & 44.7 & 22.8 & 43.9 & 54.6 \\
PANet(multi-scale)~\cite{liu2018path} & ResNeXt-101 & 47.4 & 67.2 & 51.8 & 30.1 & \textbf{51.7} & 60.0 \\
TridentNet(multi-scale)~\cite{li2019scale} & ResNet-101-DCN & \textbf{48.4} & \textbf{69.7} & \textbf{53.5} & \textbf{31.8} & 51.3 & \textbf{60.3} \\
\hline\hline
\textbf{Single-stage anchor-based:} & & & & & & & \\
SSD513~\cite{Liu2016SSD} & ResNet-101 & 31.2 & 50.4 & 33.3 & 10.2 & 34.5 & 49.8 \\
YOLOv3~\cite{redmon2018yolov3} & DarkNet-53 & 33.0 & 57.9 & 34.4 & 18.3 & 35.4 & 41.9 \\
RetinaNet800~\cite{Lin2017Focal} & ResNet-101 & 39.1 & 59.1 & 42.3 & 21.8 & 42.7 & 50.2 \\
\hline\hline
\textbf{Single-stage anchor-free:} & & & & & & & \\
ExtremeNet(single-scale)~\cite{zhou2019bottom} & Hourglass-104 & 40.2 & 55.5 & 43.2 & 20.4 & 43.2 & 53.1\\
CornerNet511(multi-scale)~\cite{law2018cornernet} & Hourglass-104 & 42.1 & 57.8 & 45.3 & 20.8 & 44.8 & 56.7 \\
FCOS~\cite{tian2019fcos} & ResNeXt-101 & 42.1 & 62.1 & 45.2 & 25.6 & 44.9 & 52.0 \\
ExtremeNet(multi-scale)~\cite{zhou2019bottom} & Hourglass-104 & 43.7 & 60.5 & 47.0 & 24.1 & 46.9 & 57.6 \\
CenterNet511(single-scale)~\cite{duan2019centernet} & Hourglass-104 & 44.9 & 62.4 & 48.1 & 25.6 & 47.4 & 57.4 \\
RPDet(single-scale)~\cite{yang2019reppoints} & ResNet-101-DCN & 45.0 & 66.1 & 49.0 & 26.6 & 48.6 & 57.5 \\
RPDet(multi-scale)~\cite{yang2019reppoints} & ResNet-101-DCN & 46.5 & \textbf{67.4} & 50.9 & \textbf{30.3} & 49.7 & 57.1 \\
CenterNet511(multi-scale)~\cite{duan2019centernet} & Hourglass-104 & 47.0 & 64.5 & 50.7 & 28.9 & 49.9 & 58.9 \\
\midrule[2pt]
CentripetalNet w.o/mask(single-scale) & Hourglass-104 & 45.8 & 63.0 & 49.3 & 25.0 & 48.2 & 58.7 \\
CentripetalNet w.o/mask(multi-scale) & Hourglass-104 & 47.8 & 65.0 & 51.5 & 28.9 & 50.2 & 59.4 \\
CentripetalNet(single-scale) & Hourglass-104 & 46.1 & 63.1 & 49.7 & 25.3 & 48.7 & 59.2 \\
CentripetalNet(multi-scale) & Hourglass-104 & \textbf{48.0} & 65.1 & \textbf{51.8} & 29.0 & \textbf{50.4} & \textbf{59.9} \\
\hline
\end{tabular}}
\end{center}
\caption{Object detection performance comparison on MS-COCO test-dev.}
\label{tbl:sota-det}
\vspace{-4mm}
\end{table*}

\section{Experiments}
\subsection{Experimental Setting}
\noindent\textbf{Dataset}
We train and validate our method on the MS-COCO 2017 dataset. 
We train our model on the train2017 split with about 115K annotated images and validate our method on the val2017 split with 5K images. We also report the performance of our model on test-dev2017 for the comparison with other detectors.

\vspace{1mm}
\noindent\textbf{Multi-task Training}
Our final objective function is
\begin{equation}
    L = L_{det} + L_{off} + {\alpha}L_{\delta} + L_{cs} + L_{mask}
\end{equation}
where $L_{det}$ and $L_{off}$ are defined as CornerNet. We set $\alpha$ to 0.05, as we find that large $\alpha$ degrades the performance of the network. As in CornerNet, we add intermediate supervision when we use Hourglass-104 as the backbone network. However, for the instance segmentation mask, we only use the feature from the last layer of the backbone to get proposals and calculate $L_{mask}$.

\vspace{1mm}
\noindent\textbf{Implementation Details}
We train our model on 16 32GB NVIDIA V100 GPUs with a batch size of 96(6 images per GPU), and we use Adam optimizer with an initial learning rate of 0.0005. To compare with other state-of-the-art models, we train our model for 210 epochs and decay the learning rate by 10 at the 180th epoch. In the ablation study, we use Hourglass-52 as the backbone and train 110 epochs, decaying the learning rate at the 90th epoch if not specified. During training, we randomly crop the input images and resize them to $511\times511$, and we also apply some usual data augmentations, such as color jitter and brightness jitter. 

During testing, we keep the resolution of input images and pad them with zeros before feeding them into the network. We use flip augmentation by default, and report both of single-scale and multi-scale test results on MS-COCO test-dev2017. To get the corners, we follow the steps of CornerNet. We firstly apply $softmax$ and $3\times3$ max pooling on the predicted corner heatmaps and select the $top100$ scored top-left corners and $top100$ bottom-right corners, then refine their locations using the predicted local offsets. Next, we can group and re-score corner pairs as described in section 3.2. In detail, we set $\mu=\frac{1}{2.1}$ for those bounding boxes with an area larger than 3500, and $\mu=\frac{1}{2.4}$ for others. Finally, we apply soft-NMS then keep the $top100$ results in the remaining bounding boxes whose scores are above 0.

\subsection{Comparison with state-of-the-art models}
\noindent\textbf{Object detection}
As shown in Table~\ref{tbl:sota-det}, CentripetalNet with Hourglass-104 as the backbone network achieves an AP of $46.1\%$ at single-scale and $48.0\%$ at multi-scale on MS-COCO test-dev 2017, which are the best performance in all anchor-free detectors. Compared to the second-best anchor-free detector, CenterNet(hourglass-104), our model achieves $1.2\%$ and $1.0\%$ AP improvement at single-scale and multi-scale separately. 
Compared to CenterNet, the improvement of CentripetalNet comes from large and medium object detection, which is just the weakness of CenterNet, as centers of large objects are more difficult to be located than those of small objects, from the perspective of probability. Compared with the two-stage detectors(without ensemble), our model is competitive as its performance is close to the state-of-the-art $48.4\% AP$ of TridentNet~\cite{li2019scale}.

Moreover, as presented in Table~\ref{tbl:ar} the AR metric of CentripetalNet outperforms all other anchor-free detectors on all sizes of objects. We suppose that the advantages of CentripetalNet's recall lie in two aspects. Firstly, the corner matching strategy based on centripetal shift can eliminate many high-scored false detections compared to CornerNet. Secondly, our corner matching strategy does not depend on the center detection, thus CentripetalNet can preserve those correct bounding boxes which are mistakenly removed in CenterNet because of the missed detection of centers.
\vspace{-1.5mm}

\begin{table}[h]
\begin{center}
\scriptsize
\begin{tabular}{|c|p{3mm}p{3.5mm}p{3.5mm}p{3.5mm}p{3.5mm}p{3.5mm}|}
\hline
Method & $AR_{1}$ & $AR_{10}$ & $AR_{100}$ & $AR_{S}$ & $AR_{M}$ &$AR_{L}$ \\ 
\hline\hline
CornerNet511-104~\cite{law2018cornernet} & 36.4 & 55.7 & 60.0 & 38.5 & 62.7 & 77.4\\
CenterNet511-104~\cite{duan2019centernet} & 37.5 & 60.3 & 64.8 & 45.1 & 68.3 & 79.7\\ \hline
CentripetalNet-104 & \textbf{37.7} & \textbf{63.9} & \textbf{68.7} & \textbf{48.8} & \textbf{71.9} & \textbf{84.0}\\ 
\hline
\end{tabular}
\end{center}
\caption{Comparison of AR metric of multi-scale test on MS-COCO test-dev2017.}
\label{tbl:ar}
\vspace{-3mm}
\end{table}

\begin{table*}[!h]
\begin{center}
\setlength{\tabcolsep}{3.8mm}{
\small
\begin{tabular}{l|l|ccc|ccc}
\hline
Method & Backbone & $AP$ & $AP_{50}$ & $AP_{75}$ & $AP_{S}$ & $AP_{M}$ & $AP_{L}$ \\
\hline\hline
PolarMask~\cite{xie2019polarmask} & ResNeXt-101 & 32.9 & 55.4 & 33.8 & 15.5 & 35.1 & 46.3 \\
ExtremeNet~\cite{zhou2019bottom}+DEXTR~\cite{maninis2018deep} & Hourglass-104 & 34.6 & 54.9 & 36.6 & 16.6 & 36.5 & 52.0 \\
Mask R-CNN~\cite{He2017Mask} & ResNeXt-101 & 37.1 & 60.0 & 39.4 & 16.9 & 39.9 & 53.5  \\
TensorMask~\cite{chen2019tensormask} & ResNet-101 & 37.1 & 59.3 & 39.4 & 17.4 & 39.1 & 51.6 \\
MaskLab+~\cite{chen2018masklab} & ResNet-101 & 37.3 & 59.8 & 39.6 & 19.1 & 40.5 & 50.6 \\
MS R-CNN~\cite{huang2019mask} & ResNeXt-101-DCN & 39.6 & 60.7 & 43.1 & 18.8 & 41.5 & 56.2 \\
HTC~\cite{chen2019hybrid} & ResNeXt-101 & 41.2 & - & - & - & - & - \\
PANet(multi-scale)~\cite{liu2018path} & ResNeXt-101 & 42.0 & 65.1 & 45.7 & 22.4 & 44.7 & 58.1 \\
\midrule[2pt]
CentripetalNet(single-scale) & Hourglass-104& 38.8 & 60.4 & 41.7 & 19.8 & 41.3 & 51.3 \\
CentripetalNet(multi-scale) & Hourglass-104& 40.2 & 62.3 & 43.1 & 22.5 & 42.6 & 52.1\\
\hline
\end{tabular}}
\end{center}
\caption{Instance segmentation performance comparison on MS-COCO test-dev.}
\label{tbl:sota-seg}
\end{table*}

\noindent\textbf{Instance segmentation}
We also report CentripetalNet's instance segmentation performance on MS-COCO test-dev 2017 for the comparison with state-of-the-art methods. As Table~\ref{tbl:sota-seg} shows, our best model achieves $38.8\%$ AP in single-scale test, while Mask R-CNN with ResNeXt-101-FPN achieves $37.5\%$ AP. 
ExtremeNet can be used for instance segmentation, with another network, DEXTR, which can convert extreme points to instance masks. However, with the same backbone, CentripetalNet achieves $4.2\%$ AP higher than ExtremeNet, and it can be trained end-to-end with the mask prediction module. Compared with the top-ranked methods, our model achieves comparable performance with a MaskAP of $40.2\%$. 

\subsection{Ablation study}

\begin{table}[]
  \begin{center}
  \scriptsize
  \begin{tabular}{|c|p{4mm}p{4mm}p{4mm}p{4mm}p{4mm}p{4mm}|}
  \hline
    & $AP$ & $AP_{50}$ & $AP_{75}$ & $AP_{S}$ & $AP_{M}$ & $AP_{L}$ \\ \hline\hline
   associative emb.(1D) & 37.3 & 53.1 & 39.0 & 17.8 & 39.4 & 50.8 \\ 
  associative emb.(2D) & 37.5 & 53.1 & 39.7 & 17.7 & 39.4 & 51.2 \\ 
  center prediction & 39.9 & 57.7 & 42.3 & 23.1 & 42.3 & 52.3 \\ 
  center regression & 40.1 & 55.8 & 42.7 & 21.0 & 42.9 & \textbf{55.6}\\ 
  centripetal shift & \textbf{40.7} & \textbf{58.0} & \textbf{42.8} & \textbf{22.4} & \textbf{43.0} & 55.4\\ \hline
  \end{tabular}
  \end{center}
  \caption{The effects of centripetal shift(without cross-star deformable convolution and mask head), compared with associative embedding, center regression and center heatmap prediction.}
  \label{tbl:abl1}
  \vspace{-2mm}
\end{table}

\noindent\textbf{Centripetal Shift}
To verify the effectiveness of our proposed centripetal shift, we conduct a series of experiments based on the corner matching methods used in previous corner-based detectors including CornerNet and CenterNet. CornerNet uses associative embedding to match corner pairs. To prove our centripetal shift's effectiveness, we replace the associative embedding of CornerNet with our centripetal shift and use our matching strategy. To be fair, we do not use the cross-star deformable convolution and expand the dimension of associative embedding to 2, the same as our centripetal shift. As shown in Table~\ref{tbl:abl1}, our method based on centripetal shift brings great performance improvement for CornerNet. As centripetal shift encodes the relationship between corner and center, direct regression to the center should have a similar effect. However, during implementation, it is sometimes impossible to apply the logarithm to the offset between the ground truth corners on heatmap and precise center locations, as the offsets sometimes may be negative because of the rounding operation when mapping the corners from original image to the heatmap. We replace the associative embedding with center regression and find that it also performs much better than CornerNet, but still worse than our centripetal shift as Table~\ref{tbl:abl1} shows. CenterNet directly predicts the center heatmap and matches the corners according to the centers and associative embedding. So we add the center prediction module to CornerNet and use the matching strategy of CenterNet, but our method still performs better, especially for large objects.

\begin{figure}[htb]
  \centering
  \includegraphics[width=1.0\linewidth]{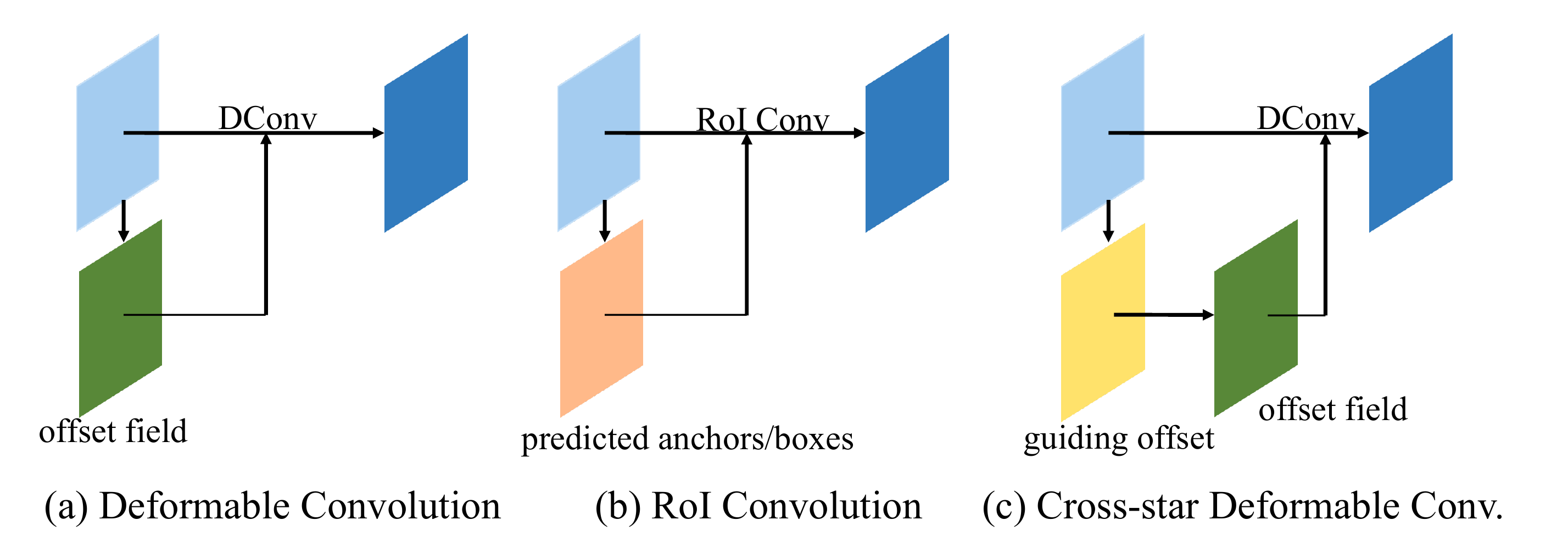}
  \caption{Different feature adaption methods. DConv means deformable convolution.}
\label{fig:fadp}
\vspace{-2mm}
\end{figure}

\begin{figure*}[htb]
  \centering
  \includegraphics[width=1\linewidth]{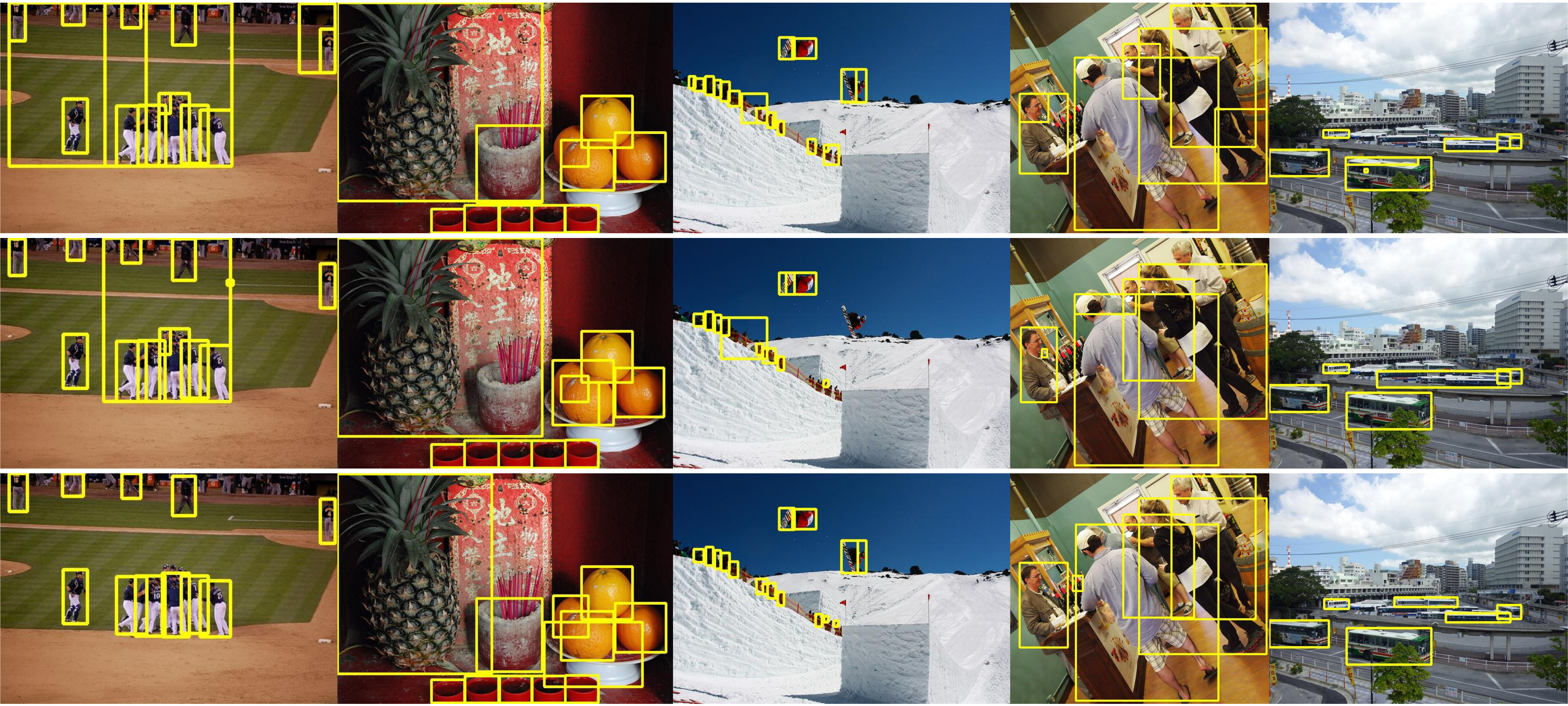}
  \caption{Above three rows show the results of CornerNet, CenterNet and CentripetalNet respectively. CornerNet and CenterNet do not perform well when the similar objects of the same category are highly concentrated. However, CentripetalNet can handle this situation.}
  \label{fig:comp}
\end{figure*}

\begin{figure}[H]
  \centering
  \includegraphics[width=1\linewidth]{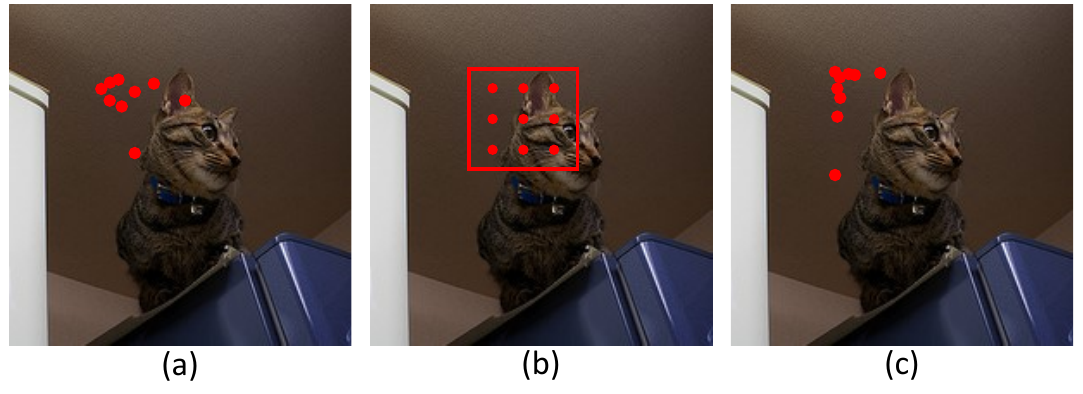}
  \caption{The sampling points of different feature adaption methods. (a) Standard deformable convolution. (b) RoI convolution. (c) Cross-star deformable convolution.}
  \label{fig:dcn_visual}
  \vspace{-2mm}
\end{figure}

\noindent\textbf{Cross-star Deformable Convolution}
Our cross-star deformable convolution is a kind of feature adaption method. Feature adaption has recently been studied in anchor-based detectors~\cite{Wang2019Region}~\cite{chen2019revisiting}, but our work is the first to discuss the topic for anchor-free detectors. Deformable convolution is usually used for feature adaption, while the main difference between different feature adaption methods is how to obtain the offset field for deformable convolution. Guided anchoring~\cite{Wang2019Region} learns the offset field from the predicted anchor shapes to align the feature with different anchor shapes at different locations in the image. AlignDet~\cite{chen2019revisiting} proposes a more precise feature adaption method, RoI convolution~\cite{chen2019revisiting}, which computes precise sampling locations for deformable convolution as shown in Figure~\ref{fig:fadp}(b). To compare RoI convolution with our feature adaption method, we regress the width and height of bounding boxes at the corners, and then we can apply RoI convolution on the feature map from corner pooling. As shown in the Table~\ref{tbl:abl2}, our method performs better than both the original deformable convolution and RoI convolution. This suggests that our cross-star deformable convolution can refine the feature for better prediction of centripetal shift. AlignDet proves that precise RoI convolution is better than learning offset field from anchor shapes. However, for our model, learning the offset field from the guiding shift performs better than RoI convolution. There are two possible reasons. First, after corner pooling, a lot of information is gathered at the border of the box instead of the inside of the box. As shown in Figure~\ref{fig:dcn_visual}, our cross-star deformable convolution tends to sample at the border of the bounding box. So it has better feature extraction ability. Second, the regression of the width and height of the bounding box is not accurate at the corner locations, so the computed sampling points of RoI convolution can not be well aligned with the ground truth. 

\begin{table}[htb]
  \begin{center}
  \scriptsize
  \begin{tabular}{|c|p{4mm}p{4mm}p{4mm}p{4mm}p{4mm}p{4mm}|}
  \hline
    & $AP$ & $AP_{50}$ & $AP_{75}$ & $AP_{S}$ & $AP_{M}$ & $AP_{L}$ \\ \hline\hline
  no feature adaption & 40.7 & 58.0 & 42.8 & 22.4 & 43.0 & 55.4 \\
   deformable conv. & 40.8 & 58.2 & 43.2 & 23.1 & 42.7 & 54.9 \\
  RoI conv. & 41.1 & 58.5 & 43.4 & 22.9 & 43.4 & 55.5 \\
  cross-star deformable conv. & \textbf{41.5} & \textbf{58.7} & \textbf{44.4} & \textbf{23.3} & \textbf{44.1} & \textbf{55.7} \\ \hline
  \end{tabular}
  \end{center}
  \caption{Comparison of different feature adaption methods. Base model is CentripetalNet without feature adaption and mask head, then we add different feature adaption modules separately.}
  \label{tbl:abl2}
  \vspace{-4mm}
\end{table}

\begin{figure}
    \centering
    \includegraphics[width=0.95\linewidth]{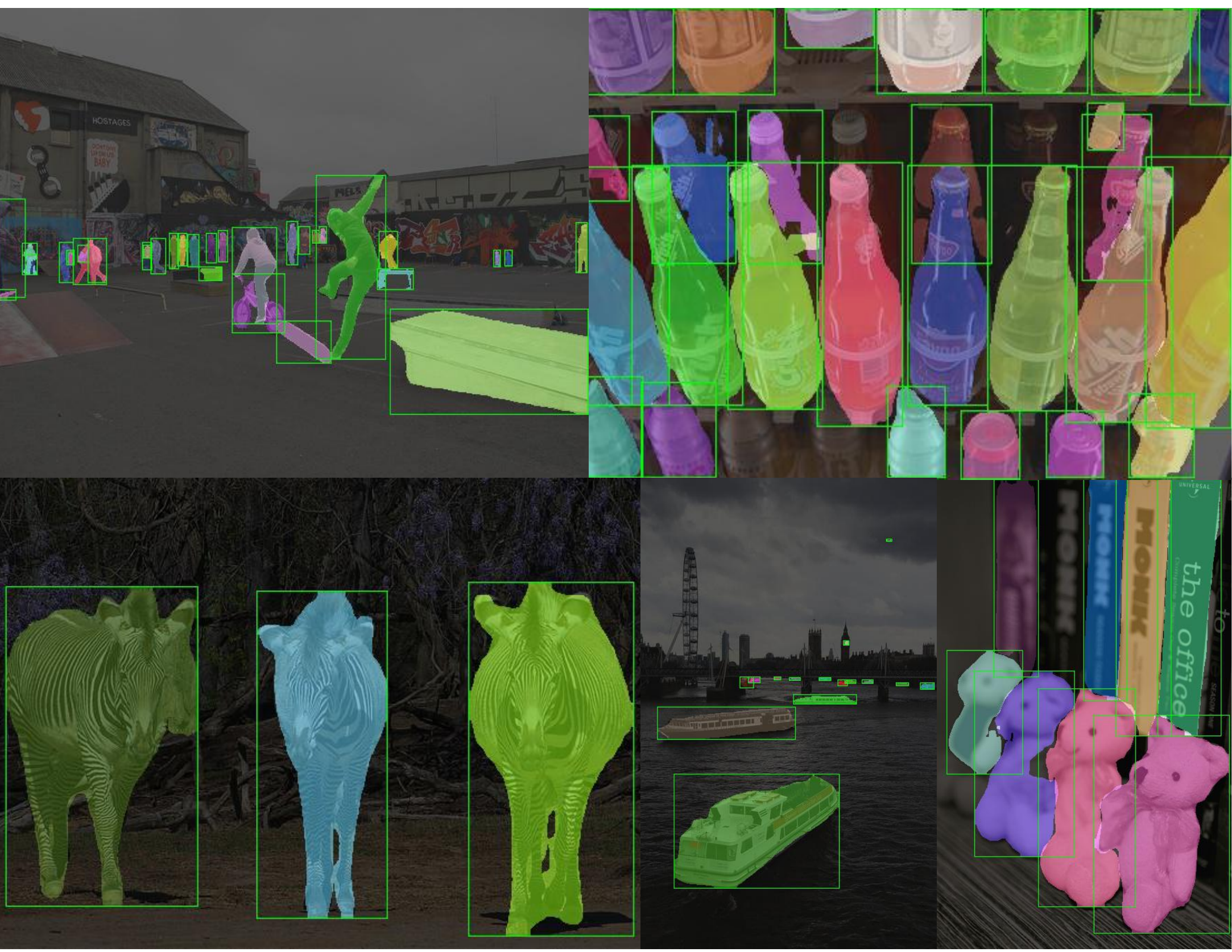}
    \caption{CentripetalNet instance segmentation results on MS-COCO val2017.}
    \label{fig:mask}
\end{figure}

\vspace{1mm}
\noindent\textbf{Instance Segmentation Module}
Many works~\cite{He2017Mask,fu2019retinamask:} have proved that the instance segmentation task can improve the performance of anchor-based detectors. Hence we add a mask prediction module as described in section 3.3. 
As Table~\ref{tbl:abl3} shows, multi-task learning improves our model's $AP_{bbox}$ by 0.3$ \% $, when training 110 epochs. If we train CentripetalNet with 210 epochs, the improvement becomes 0.4$ \% $. 
We find that mask head does not improve the performance of CornerNet at all. This result shows that this multi-task learning has almost little influence on the corner prediction and associative embedding prediction, but benefits the prediction of our centripetal shift. As shown in Figure~\ref{fig:mask}, CentripetalNet can generate fine segmentation masks.

\begin{table}
  \begin{center}
  \scriptsize
  \begin{tabular}{|c|c|p{2mm}p{3.5mm}p{3.5mm}p{3.5mm}p{3.5mm}p{3.5mm}|}
  \hline
    & epoch & $AP$ & $AP_{50}$ & $AP_{75}$ & $AP_{S}$ & $AP_{M}$ & $AP_{L}$ \\ \hline \hline
  CornerNet & 110 & 37.3 & 53.1 & 39.0 & 17.8 & 39.4 & 50.8 \\
   CornerNet w/mask& 110 & 37.3 & 53.0 & 39.5 & 18.3 & 39.2 & 50.7 \\
  CentripetalNet w.o/mask  & 110 & 41.5 & 58.7 & 44.4 & 23.3 & 44.1 & 55.7 \\
  CentripetalNet & 110 & 41.8 & 58.9 & 44.5 & 23.0 & 44.1 & 56.7 \\ \hline\hline
  CentripetalNet w.o/mask  & 210 & 41.7 & \textbf{59.0} & 44.4 & 23.3 & 44.4 & 56.1 \\
  CentripetalNet & 210 & \textbf{42.1} & 58.7 & \textbf{44.9} & \textbf{23.7} & \textbf{44.5} & \textbf{56.8} \\ \hline
  \end{tabular}
  \end{center}
  \caption{The effect of mask prediction module on CornerNet and CentripetalNet, both with Hourglass-52 as backbone.}
  \label{tbl:abl3}
  \vspace{-3mm}
  \end{table}
  
\vspace{-0mm}
\subsection{Qualitative analysis}
\vspace{-0mm}
As Figure~\ref{fig:comp} shows, CentripetalNet successfully removes the wrong corner pairs in CornerNet. Compared to CenterNet, CentripetalNet has two advantages. Firstly, CentripetalNet does not rely on center detections, so it can keep the correct predicted bounding boxes, which are incorrectly deleted in CenterNet due to the missed detection of centers. Secondly, CenterNet cannot handle the situations in which the center of an object is in the central region of a box composed of the corners of another two objects. This situation usually occurs in a dense situation, such as the crowd.

\section{Conclusion}
In this work, we introduce simple yet effective centripetal shift to solve the corner matching problem in recent anchor-free detectors. Our method establishes the relationship between corners through positional and geometric information and overcomes the ambiguity of associative embedding caused by similar appearance. Besides, we equip our detector with an instance segmentation module and firstly conduct end-to-end instance segmentation using the anchor-free detector. Finally, the state-of-the-art performance on MS-COCO proves the strength of our method.


{\small
\bibliographystyle{ieee_fullname}
\bibliography{egbib}
}

\end{document}